\documentclass[11pt,a4paper]{article}
\usepackage[hyperref]{emnlp2020}
\usepackage{times}
\usepackage{latexsym}

\usepackage{amsmath}

\usepackage{microtype}
\usepackage{subfigure}
\usepackage{mathtools}
\usepackage{pgfplots}
\usepackage{multirow}
\usepackage{graphicx}
\usetikzlibrary{calc}
\usepackage{ctable}
\DeclareMathOperator*{\argmax}{arg\,max}

\usepackage{amssymb}

\aclfinalcopy 

\newif\ifcomments
\commentstrue 
\ifcomments
    \providecommand{\jiangming}[1]{{\protect\color{red}{[J: #1]}}}
    \providecommand{\matt}[1]{{\protect\color{teal}{[M: #1]}}}
\else
    \providecommand{\matt}[1]{}
    \providecommand{\jiangming}[1]{}
\fi

\title{Multi-Step Inference for Reasoning Over Paragraphs}

\author{Jiangming Liu$^{\spadesuit}$\thanks{~~Work done during an internship at Allen Institute for Artificial Intelligence.},~Matt Gardner$^{\clubsuit}$,~Shay B. Cohen$^{\spadesuit}$,~Mirella Lapata$^{\spadesuit}$ \\
  $^{\spadesuit}$ ILCC, University of Edinburgh \\
  $^{\clubsuit}$ Allen Institute for AI \\
  \texttt{jiangming.liu@ed.ac.uk},~
  \texttt{mattg@allenai.org},~\\
  \texttt{\{scohen,mlap\}@inf.ed.ac.uk} \\}

\date{}

\begin{document}
\maketitle
\begin{abstract}
Complex reasoning over text requires understanding and chaining together free-form predicates and logical connectives. Prior work has largely tried to do this either symbolically or with black-box transformers.  We present a middle ground between these two extremes: a compositional model reminiscent of neural module networks that can perform chained logical reasoning.  This model first finds relevant sentences in the context and then chains them together using neural modules.  Our model gives significant performance improvements (up to 29\% relative error reduction when combined with a reranker) on ROPES, a recently-introduced complex reasoning dataset.
\end{abstract}

\section{Introduction}

Performing chained inference over natural language text is a long-standing goal in artificial intelligence \cite{grosz1986readings,reddy2003three}.  This kind of inference requires understanding how natural language statements fit together in a way that permits drawing conclusions.  This is very challenging without a formal model of the semantics underlying the text, and when polarity needs to be tracked across many statements.

For instance, consider the example in Figure 1 
from ROPES \cite{lin2019reasoning}, a recently released reading comprehension dataset that requires applying information contained in a background paragraph to a new situation.  To answer the question, one must associate each category of flowers with a polarity for having brightly colored petals, which must be done by going through the information about pollinators given in the situation and linking it to what was said about pollinators and brightly colored petals in the background paragraph, along with tracking the polarity of those statements.
\begin{figure}[!ht]
    \centering
    \subfigure[]
    {
    \begin{tabular}{p{7.3cm}}
    \hline
    \textbf{Background:} {\small Scientists think that the earliest flowers attracted \textit{\textcolor{blue}{insects and other animals, which spread pollen}} from flower to flower. \textit{\textcolor{blue}{This greatly increased the efficiency of fertilization}} \textit{\textcolor{blue}{over wind-spread pollen}}, which might or might not actually land on another flower. \textit{\textcolor{blue}{To take better advantage of} \fbox{\textcolor{blue}{this animal labor}}, \textcolor{blue}{plants evolved traits such as} \fbox{\textcolor{blue}{brightly colored petals to attract pollinators.}}} In exchange for pollination, flowers gave the pollinators nectar.}\\
    \hline
    \hline
   \textbf{Situation:} {\small Last week, John visited the national park near his city. He saw many flowers. His guide explained him that there are two categories of flowers, category A and category B. \textit{\textcolor{red}{Category A flowers spread pollen via wind, and category B flowers spread pollen via animals.}}} \\
    \hline
    \hline
    \textbf{Question:} {\small Which category of flowers would be \textit{\textcolor{orange}{more likely to have brightly colored petals}}?} \\
    \textbf{Answer:} {\small category B} \\
    \hline
    \end{tabular}
    }
    \subfigure[]
    {
    \begin{tikzpicture}[font=\small,scale=0.8]
    \tikzset{
        passage/.style={anchor=base,outer sep=1pt}
        }
    \node[passage] (select) at (0, 0) {\textsc{Select}};
    \node[passage] (select) at (0, -1) {\textsc{Chain}};
    \node[passage] (select) at (0, -2) {\textsc{Chain}};
    \node[passage] (select) at (0, -3) {\textsc{Predict}};
    \node[passage] (btext) at (2, 1) {background};
    \node[passage] (qtext) at (4, 1) {question};
    \node[passage] (stext) at (6, 1) {situation};
    
    \node[circle, fill=blue!80, minimum height=0.5cm, minimum width=0.5cm, outer sep=1pt] (bselect) at (2, 0) {}; 
    
    \node[circle, fill=orange!80, minimum height=0.5cm, minimum width=0.5cm, outer sep=1pt] (qselect) at (4, 0) {};
    
    \node[rectangle, fill=white!100, draw=black!100,minimum height=0.5cm, minimum width=0.5cm, outer sep=1pt] (bchain) at (3, -1) {};
    
    \node[circle, fill=red!80, minimum height=0.5cm, minimum width=0.5cm, outer sep=1pt] (schain) at (4, -2) {};
    
    \node[circle, fill=blue!80, minimum height=0.3cm, minimum width=0.3cm, inner sep=0pt, outer sep=1pt] (bselect_p) at (2.5, -3) {};
    \node[rectangle, fill=white!100, draw=black!100,minimum height=0.3cm, minimum width=0.3cm, outer sep=1pt] (bchain_p) at (3, -3) {};
    \node[circle, fill=orange!80, minimum height=0.3cm, minimum width=0.3cm, outer sep=1pt] (qselect_p) at (3.5, -3) {};
    \node[circle, fill=red!80, minimum height=0.3cm, minimum width=0.3cm, outer sep=1pt] (schain_p) at (4, -3) {};
    \path[dashed]
        (bselect) edge (bselect_p)
        (qselect) edge (qselect_p)
        (bchain) edge (bchain_p)
        (schain) edge (schain_p);
    \node[passage] (ans) at (3.25, -4.5) {category B};
    \path[->]
        (3.25,-3.5) edge (ans.north);
    \path[->]
        (btext) edge (bselect)
        (qtext) edge (qselect)
        (btext) edge [bend left=30] (bchain)
        (qselect) edge (bchain)
        (bselect) edge [bend right=50] (schain)
        (bchain) edge (schain)
        (stext) edge [bend left=20] (schain);
    
    \draw[dashed] (-1, 0.5) -- (7, 0.5);
    \draw[dashed] (-1, -0.5) -- (7, -0.5);
    \draw[dashed] (-1, -1.5) -- (7, -1.5);
    \draw[dashed] (-1, -2.5) -- (7, -2.5);
    \end{tikzpicture}
    }
    \caption{(a) An example in ROPES; (b) the chained reasoning that our model performs on the example. The model first (softly) selects relevant parts of the background and question, then successively chains them, making a prediction after including the situation in the chaining.}
    \label{example}
\end{figure}

Prior work addressing this problem has largely either used symbolic reasoning, such as markov logic networks \cite{khot2015exploring} and integer linear programming \cite{khashabi2016question}, or black-box neural networks \cite{jiang-etal-2019-explore,jiang-bansal-2019-self}.  Symbolic methods give some measure of interpretability and the ability to handle logical operators to track polarity, but they are brittle, unable to handle the variability of language.  Neural networks often perform better on practical datasets, as they are more robust to paraphrase, but they lack any explicit notion of reasoning and are hard to interpret.

We present a model that is a middle ground between these two approaches: a compositional model reminiscent of neural module networks that can perform chained logical reasoning. The proposed model is able to understand and chain together free-form predicates and logical connectives.
The proposed model is inspired by neural module networks (NMNs), which were proposed for visual question answering~\cite{nmn1,andreas-etal-2016-learning}. NMNs assemble a network from a collection of specialized modules where each module performs some learnable function, such as locating a question word in an image, or recognizing relationships between objects in the image. The modules are composed together specific to what is asked in the question, then executed to obtain an answer. We design general modules that are targeted at the reasoning necessary for ROPES and compose them together to answer questions.

We design three kinds of basic modules to learn the neuro-symbolic multi-step inference over questions, situations, and background passages.
The first module is called \textsc{Select}, which determines which information (in the form of spans) is essential to the question; the second module is called \textsc{Chain}, which captures the interaction from multiple statements; the last one is called \textsc{Predict}, which assigns confidence scores to potential answers.
The three basic modules can be instantiated separately and freely combined. 

In this paper, we investigate one possible combination as our multi-step inference on ROPES. The results show that with the multi-step inference, the model achieves significant performance improvement. Furthermore, when combined with a reranking architecture, the model achieves a relative error reduction of 29\% and 8\% on the dev and test sets in the ROPES benchmark.  As ROPES is a relatively new benchmark, we also present some analysis of the data, showing that the official dev set is likely better treated as an in-domain test, while the official test set is more of an out-of-domain test set.\footnote{Model code is available at \url{https://github.com/LeonCrashCode/allennlp/blob/transf-exp1/allennlp/models/transformer_mc/roberta_models.py}}

\section{Model}
We first describe the baseline system, a typical QA span extractor built on \textsc{RoBERTa}~\cite{liu2019roberta}, and then present the proposed system with multi-step inference. Furthermore, we introduce a reranker with multi-step inference given the output of the baseline system.

Following the standard usage of \textsc{RoBERTa}, we concatenate the background, the situation and question with two special determiners [S:] and [Q:] to be a long passage $P =$ [CLS] $B$ [S:] $S$ [SEP] [SEP] [Q:] $Q$ [SEP], where the background $B$ and situation $S$ are regarded as the first segment and the question $Q$ is the second segment, and [CLS] and [SEP] are the reserved tokens in \textsc{RoBERTa}.

\subsection{Baseline}
Our baseline system is a span extractor built on the top of \textsc{RoBERTa}.
Given the passage representations from \textsc{RoBERTa} $P_{\textsc{RoBERTa}} = [x_0, ..., x_{n-1}]$, two scores are generated for each token by span scorer, showing the chance to be the start and the end of the answer span:
\begin{equation*}
    \bar{S}, \bar{E} = \text{QA\_score}(P_{\textsc{RoBERTa}}),
\end{equation*}
where $\bar{S} = [\bar{s}, \bar{s}_1, ...\bar{s}_{n-1}$] and $\bar{E} = [\bar{e}_0, \bar{e}_1, ..., \bar{e}_{n-1}] ~(0 \leq k < n)$\footnote{The answer spans always appear in the situation and question passage, so we mask the scores for the background passage.}are the scores of the start and the end of answer spans, respectively. QA\_score$(\cdot):\mathbb{R}^{d_x} \Rightarrow \mathbb{R}^2$ is a linear function, where $d_x$ is the output dimension of \textsc{RoBERTa}.
The span with highest start and end scores is extracted as the answer by span extractor:
\begin{equation*}
\begin{split}
    [s_0, s_1, ..., s_n] & = \textsc{Softmax}([\bar{s}_0, \bar{s}_1, ..., \bar{s}_n]) \\
    [e_0, e_1, ..., e_n] & = \textsc{softmax}([\bar{e}_0, \bar{e}_1, ..., \bar{e}_n]) \\
    i^*, j^* = \argmax_{i,j} &~s_i + e_j ~~~ (0 \leq i \leq j < n),
\end{split}
\end{equation*}
where the span$_{i^*,j^*}$ is the answer. 

\subsection{Multi-Step Inference for ROPES}
Instead of a simple span prediction head on top of \textsc{RoBERTa}, our proposed multi-step inference model uses a series of neural modules targeted at chained inference.  Like the baseline, our model begins with encoded \textsc{RoBERTa} passage representations $P_{\textsc{RoBERTa}}$, but replaces the QA\_score function with a MS-Inference function, which similarly outputs a span start and end score for each encoded token $x_k$:
\begin{equation*}
    \bar{S}, \bar{E} = \text{MS-Inference}(P_{\textsc{RoBERTa}})
\end{equation*}

The MS-Inference($\cdot$) function consists of several modules. These modules \textsc{Select} relevant information from parts of the passage, \textsc{Chain} the selected text together, then \textsc{Predict} the answer to the question given the result of the chaining. These modules are applied on $P_{\textsc{RoBERTa}}$ which is decomposed into  $\textit{B}_{\textsc{RoBERTa}}, \textit{S}_{\textsc{RoBERTa}}, \textit{Q}_{\textsc{RoBERTa}}$, denoting the token representations of \textsc{RoBERTa} for the background, the situation and the question, respectively.

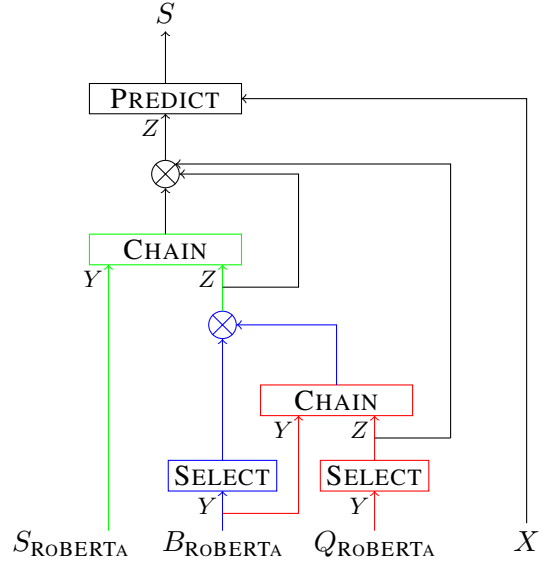
\begin{figure}[t]
    \centering
    \begin{tikzpicture}[scale=1]
    \tikzset{
        passage/.style={anchor=base,inner sep=0pt},
        interface/.style={anchor=base,inner sep=0pt, font=\small},
        smodule/.style = {rectangle, fill=white!100, draw=black!100, minimum width=1cm,outer sep=0pt,inner sep=2pt},
        cmodule/.style = {rectangle, fill=white!100, draw=black!100, minimum width=2cm,outer sep=0pt,inner sep=2pt},
        concate/.style = {circle, draw=black!100}
        }
    \node[passage] (si) at (-2, -1) {$S_{\textsc{RoBERTa}}$};
    \node[passage] (bg) at (0, -1) {$B_{\textsc{RoBERTa}}$};
    \node[passage] (qu) at (2, -1) {$Q_{\textsc{RoBERTa}}$};
    \node[passage] (cands) at (4, -1) {$X$};
    
    \node[smodule,draw=red] (qsolo) at (2, 0) {\textsc{Select}};
    \node[interface] (qsolo_input) at ([xshift=-0.2cm,yshift=-0.3cm]qsolo.south) {$Y$};
    \node[smodule,draw=blue] (bsolo) at (0, 0) {\textsc{Select}};
    \node[interface] (bsolo_input) at ([xshift=-0.2cm,yshift=-0.3cm]bsolo.south) {$Y$};
    \path[->,red]
        (qu) edge (qsolo);
    \path[->,blue]
        (bg) edge (bsolo);
    \node[cmodule, draw=red] (bqcon) at (1.5, 1) {\textsc{Chain}};
    \node[interface] (bqcon_input1) at ([xshift=-0.7cm,yshift=-0.3cm]bqcon.south) {$Y$};
    \node[interface] (bqcon_input2) at ([xshift=0.3cm,yshift=-0.3cm]bqcon.south) {$Z$};
    \draw[->,red] (0,-0.5) -- (1, -0.5) -- ([xshift=-0.5cm]bqcon.south);
    \draw[->,red] (qsolo.north) -- ([xshift=0.5cm]bqcon.south);
    
    \node[concate, blue] (bqx) at (0, 2) {};
    \draw[-,blue] (bqx.south west) -- (bqx.north east) (bqx.south east) -- (bqx.north west);
    \draw[->,blue] (bsolo) -- (bqx.south);
    
    \draw[->,blue]
        let \p{B}=(bqx.east), \p{A}=(bqcon.north) in
        (\x{A}, \y{A}) -- (\x{A}, \y{B}) -- (\x{B}, \y{B});
    \node[cmodule, draw=green] (con) at (-0.75, 3) {\textsc{Chain}};
    \node[interface] (con_input1) at ([xshift=-0.95cm,yshift=-0.3cm]con.south) {$Y$};
    \node[interface] (con_input2) at ([xshift=0.55cm,yshift=-0.3cm]con.south) {$Z$};
    \draw[->, green]
        (bqx.north) -- ([xshift=0.75cm]con.south);
    \draw[->, green]
        ([xshift=0.5cm]si.north) -- ([xshift=-0.75cm]con.south);
    \node[cmodule] (score) at (-0.75, 5) {\textsc{Predict}};
    \node[interface] (score_input) at ([xshift=-0.2cm,yshift=-0.3cm]score.south) {$Z$};
    \node[concate] (all) at (-0.75, 4) {};
    \draw[-] (all.south west) -- (all.north east) (all.south east) -- (all.north west);
    \draw[->] 
        let \p{B}=(score.east), \p{A}=(cands.north) in
        ([yshift=0.1cm]cands.north) -- (\x{A}, \y{B}) -- (\x{B}, \y{B});
    \draw[-] (2, 0.5) -- (3, 0.5);
    \draw[->]
        let \p{B}=(all.north east), \p{A}=(3, 0.5) in
        (\x{A}, \y{A}) -- (\x{A}, \y{B}) -- (\x{B}, \y{B});
    \draw[->]
        (con.north) -- (all.south);
    \draw[->]
        (all.north) -- (score.south);
    \draw[->]
        (0, 2.5) -- (1, 2.5) -- (1, 4) -- (all.east);
    \node[passage] (scores) at (-0.75, 6) {$S$};
    \draw[->]
        (score.north) -- ([yshift=-0.1cm]scores.south);
    \end{tikzpicture}
    \caption{Multi-step inference model, where $\otimes$ is the operation to collect multiple vectors as a list, $Z,Y$ are the interfaces of the modules, and $X$ is a token representation to be scored as start/end of the answer span in the QA systems, or a candidate span representation to be scored in the reranking systems.}
    \label{multistepinference}
\end{figure}

As most of the questions in ROPES require the same basic reasoning steps, we use a fixed combination of these modules to answer every question, instead of trying to predict the module layout for each question, as was done in prior work~\cite{Hu2017LearningTR}.  This combination is shown in Figure~\ref{multistepinference}: we \textsc{Select} important parts of the question passage, and \textsc{Chain} them with the background passage to find a likely part of the background that supports answering the question (marked as red). Then we \textsc{Select} important parts of the background passage, which are combined with previous results that we have (marked as blue), and we \textsc{Chain} the combined information to find relevant parts of the situation passage (marked as green), and finally \textsc{Predict} an answer (marked as black), which is most often found in the situation text.  The intuition for how these modules work together to piece together the information necessary to answer the question is shown in Figure~\ref{example}.
The actual operations performed by each of these modules is described below.

\paragraph{\textsc{Select}}
The select module, i.e.~$z = \textsc{Select}(Y)$, where $Y\in \mathbb{R}^{n\times d_x}$ and $z\in \mathbb{R}^{d_x}$, aims to find the important parts of its input and summarize in a single vector. It first uses a learned linear scoring function, $f(\cdot):\mathbb{R}^{ d_{x}}\Rightarrow\mathbb{R}$, to determine which parts of its input are most important, then converts those scores into a probability distribution using a \textsc{Softmax} operation, and computes a weighted sum of the inputs:
\begin{equation*}
\begin{split}
    W & = f(Y) \\
    A &= \textsc{Softmax}(W) \\
    z &= A^\mathrm{T}Y,
\end{split}
\end{equation*}

\paragraph{\textsc{Chain}}
The chain module, i.e.~$z = \textsc{Chain}(Y,Z)$, computes the interaction between an input matrix $Y$ and a list of the input vectors $Z = [z_0, z_1, ..., z_{l-1}]$, where $Y \in \mathbb{R}^{n\times d_x}$, $z_k \in \mathbb{R}^{d_k}$ and $d_k$ is the dimension of the $k$th input vector~$(0 \leq k < l)$, and again outputs a summary vector of this interaction $z \in \mathbb{R}^{d_x}$.  Intuitively, this module is supposed to chain together the inputs $Y$ and $Z$ and return a summary of the result.  This is done with the following operations:
\begin{equation*}
\begin{split}
    z' &= g([z_0; z_1; ...; z_{l-1}]) \\ 
    z &= \textsc{Attention}(z', Y, Y),
\end{split}
\end{equation*}
where $g(\cdot)~:~ \mathbb{R}^{(d_0+d_1+...+d_{l-1})} \Rightarrow \mathbb{R}^{d_x}$ is a linear function,~$;$~is the concatenation, and attention($\cdot$) is instantiated with the multi-head attention:
\begin{equation*}
    \textsc{Attention}(z', Y, Y) = [\text{att}_1; \text{att}_2; ...; \text{att}_h]W^O 
\end{equation*}
\begin{equation*}
\begin{split}
    \text{att}_k &= \text{att}(z'W^Q_k, YW^{K}_k, YW^{V}_k) \\
    \text{att}(Q, K, V) &= \textsc{SoftMax}(\frac{QK^{\mathrm{T}}}{\sqrt{d_x}})V, \\
\end{split}
\end{equation*}
where $W^O, W^Q_k, W^K_k$ and $W^V_k$ are trainable parameters.

\paragraph{\textsc{Predict}}
The predict module, i.e.~$S = \textsc{Predict}(Z, X)$, takes the list of output vectors $Z = [z_0, z_1, ..., z_{m-1}]$ from previous modules, where $z_k \in \mathbb{R}^{d_k} (0 \leq k < m)$ and $m$ is the number of previous modules, and the candidates $X = [x_0, x_1, ..., x_{n-1}]$, where $x_k \in \mathbb{R}^d_x$ and $n$ is the number of candidates, and produces scores for the candidates.  In our base model, each candidate is a token in the situation or question, and the score is a pair of numbers representing span start and end probabilities for that token.  When we use this module in a re-ranker (Section~\ref{reranker}), the candidates $X$ are already encoded spans, and so we produce just one number for each span.  The \textsc{Predict} module simply uses a linear scoring function on the concatenation of its inputs:
\begin{equation*}
\begin{split}
    S = & [s_0, s_1, ..., s_{n-1}] \\
    s_k = & \textsc{Score}([z_0; z_1; ...; z_l; x_k]),
\end{split}
\end{equation*}
where $\textsc{Score}(\cdot): \mathbb{R}^{1 \times (d_0+d_1+...+d_{m-1}+d_{x})} \Rightarrow \mathbb{R}^r$ is a linear function, $;$ is the concatenation and $r=2$ if the module is used to extract spans, while $r=1$ if the module is used to score candidates for the reranker.

\paragraph{Full model}

Our full model combines these modules in the following way to compute span start and end scores for each token (depicted graphically in Figure~\ref{multistepinference}):
\begin{equation*}
\begin{split}
    S_{\textsc{RoBERTa}},& B_{\textsc{RoBERTa}}, Q_{\textsc{RoBERTa}} = P_{\textsc{RoBERTa}} \\
    X &= [S_{\textsc{RoBERTa}}; Q_{\textsc{RoBERTa}}] \\
    z_0 &= \textsc{Select}(Q_{\textsc{RoBERTa}}) \\
    z_1 &= \textsc{Select}(B_{\textsc{RoBERTa}}) \\
    z_2 &= \textsc{Chain}(B_{\textsc{RoBERTa}}, [z_0]) \\
    z_3 &= \textsc{Chain}(S_{\textsc{RoBERTa}}, [z_1, z_2]) \\
    \bar{S}, \bar{E} = S &= \textsc{Predict}([z_0, z_1, z_2, z_3], X) \\
\end{split}
\end{equation*}

\subsection{Multi-Step Reranker}
\label{reranker}
Most questions in ROPES have only two or three reasonable candidate answers (in Figure~\ref{example} these are ``category A'' and ``category B''), and we find that the baseline model is able to reliably find these answers, though it has a hard time selecting between them.  This suggests that a reranker that only focuses on deciding which of the candidates is correct could be effective.  To do this, we take the top $c$ spans output by the baseline system and score these candidates directly using our MS-Inference model instead of producing span start and end scores for each input token.

\paragraph{Scoring spans instead of tokens} To feed the candidate spans into our multi-step inference model, we represent each span as a single vector by concatenating its endpoint tokens: $x_{(i,j)} = [x_i; x_j]$. We take all $c$ candidates and concatenate them together as $X$, instead of $X = [S; Q]$ as is done in our base model.  Similarly, $\textsc{Predict}(Z,X)$ outputs a single score $\bar{O}$ per candidate instead of a pair of start and end probabilities.

\paragraph{Ensemble}
We additionally use an ensemble strategy for the reranker. We train several rerankers and build a voting system where each reranker makes a vote for the candidate to be the best answer. The candidate with the most votes is chosen the best answer through the voting system.


\begin{table*}[!ht]
    \centering
    \begin{tabular}{c|p{13cm}}
    \specialrule{.1em}{.05em}{.05em} 
       Type  &  Passage \\
       \hline
        \multirow{2}{*}{NP} &  ...The child poured two spoonfuls of sugar into cup A and three spoonfuls of sugar into \textbf{\textcolor{red}{cup B}}... Which cup has a higher concentration of sugar ?  \\
        \cline{2-2} 
        & ...They labeled it as \textbf{\textcolor{red}{plant B}} . They wanted to find out what makes a plant drought-resistant... In which plant there would be more water loss ?\\
        \hline
        \multirow{2}{*}{VP} & ...In test B he used higher concentration of reactants. Now, he needs to know about the science... Would test B \textbf{\textcolor{red}{increase}} or decrease the frequency...\\
        \cline{2-2}
        & ...induced higher respiration rate in sample A. Then he induced no respiration rate in sample B... make their own glucose or \textbf{\textcolor{red}{acquire it from other organisms}} ?\\
        \hline
        \multirow{2}{*}{ADJP} & ... patient A and patient B. John found out that patient A had more LDL, but patient B had more HDL... B have higher or \textbf{\textcolor{red}{lower}} risk of heart attack than patient A?\\
        \cline{2-2}
        & ...visible light. He noted microwaves as case A, infrared as case B, and visible light as case C...Would case A have \textbf{\textcolor{red}{longer}} or shorter wavelengths than case B?\\
        \hline
        \multirow{2}{*}{ADVP} &...Sample A was a strong acid, and sample B was a weak acid. David needed to ...sample A lose a proton less or \textbf{\textcolor{red}{more easily}} than sample B?\\
        \cline{2-2}
        &...There is only one ice cube left so she takes it out and sets it in the glass on the table. She then refills...in the ice cube moving closer together or \textbf{\textcolor{red}{farther apart}} ?\\
        \hline
        \multirow{2}{*}{Others} & ...Their mother takes them to see a doctor and to have their testosterone tested. The tests reveal that...Will Jimothy finish his growth spurt before or \textbf{\textcolor{red}{after}} Dwight?\\
        \cline{2-2}
        & ...He cut down on how much he eats every day and monitors his calorie intake, making sure that he is...Given Greg's BMI us 41, is he considered obese, \textbf{\textcolor{red}{yes}} or no?\\
    \specialrule{.1em}{.05em}{.05em} 
    \end{tabular}
    \caption{The examples in ROPES, where the bold red spans are answers.}
    \label{example}
\end{table*}

\section{Data bias in ROPES}
\label{bias}
We experiment with ROPES~\cite{lin2019reasoning}, a recently proposed dataset which focuses on complex reasoning over paragraphs for document comprehension.  We noticed a very severe drop in performance between the ROPES dev and test sets during initial experiments, and we performed an analysis of the data to figure out the cause.  ROPES used an annotator split to separate the train, dev, and test sets in order to avoid annotator bias~\cite{geva2019we}, but we discovered that this led to a large distributional shift between train/dev and test, which we explore in this section. In light of this analysis, we recommend treating the dev set as an in-domain test set, and the original test set as an out-of-domain test.

\paragraph{Answer types} Our analysis is based on looking at the syntactic category of the answer phrase.  We use the syntactic parser of \newcite{Kitaev-2018-SelfAttentive} to obtain constituent trees for the passages in ROPES. We take the constituent label of the lowest subtree that covers the answer span\footnote{ The passages could have more than one span that matches the answer; we use the last occurrence of the answer span for our analysis.} as the answer type.

The four most frequent answer types in ROPES are noun phrase (NP), verb phrase (VP), adjective phrase (ADJP) and adverb phrase (ADVP). Table~\ref{example} shows examples for each type. Most NP answers come from the situation, while the other answer types typically come from the question.

\paragraph{Bias}
The distribution of answer types in the train/dev/test sets of ROPES are shown in Table~\ref{type}.  We found that the distribution in the train set is similar to development set, where most of the answers are NPs (85\%), with ADJP being the second most frequent. However, the test set has a very different distribution over answer types, where less than half of the answers NPs, and there are more VPs, ADJPs, ADVPs, and other types. 

This distributional shift over answer types between train/dev and test raises challenges for reading comprehension systems; to perform well on test, the model must predict a significant number of answers from the question instead of from the situation, which only rarely happens in the training data.  Given this distributional shift, it seems fair to characterize the official test as somewhat out-of-domain for the training data.

\begin{table}[]
    \centering
    \begin{tabular}{cccc}
    \toprule 
      Answer type &  Train & Dev & Test \\
    \midrule
        NP & 84.17 & 85.19 & 47.19 \\
        VP & 3.35 & 1.24 & 17.37 \\
        ADJP & 9.20 & 10.25 & 19.36 \\
        ADVP & 2.50 & 3.32 & 10.23 \\
        Others & 0.78 & 0.00 & 5.85 \\
    \bottomrule 
    \end{tabular}
    \caption{The percentage (\%) of question types in ROPES.}
    \label{type}
\end{table}

\section{Experiments}

In this section, we evaluate the performance of our proposed model relative to baselines on ROPES.

\begin{table}[!ht]
    \centering
    \begin{tabular}{lccc}
    \toprule 
         &  Train & Dev & Test \\
    \midrule
    \# of backgrounds & 513 & 51& 171 \\
    \# of situations & 1,409 & 203 & 300 \\
    \# of questions & 10,924 & 1,688 & 1,710 \\
    \bottomrule
    \end{tabular}
    \caption{The ROPES dataset}
    \label{dataset}
\end{table}

\subsection{Settings}
\paragraph{Data}
We use the 10,924 questions as our training set, and 1,688 questions as dev set and 1,710 questions as test set, where each question has only one answer, which is a span from either the situation or the question. Table~\ref{dataset} shows the statistics on the ROPES benchmark.
Due to the severe distributional shift between dev and test (described in Section~\ref{bias}), we additionally set up an experiment using the dev set as an in-domain test set, by partitioning the training set into train (9,824 questions) and train-dev (1,100 questions).
\vspace{-0.2cm}
\paragraph{Training}
Following the settings of prior work \cite{lin2019reasoning}, we fine-tune the \textsc{RoBERTa-large} pre-trained transformer. The hidden sizes of all layers are set to 1024 which is the same to the output dimension of \textsc{RoBERTa-large}, and the number of heads on multi-step attentions is 8. All the models share the same hyperparameters that are shown in Table \ref{hyper}.\footnote{The hyperparamters are manually tuned according to the performance on dev dataset.}

\begin{table}[]
    \centering
    \begin{tabular}{l|c}
    \specialrule{.1em}{.05em}{.05em}
        parameter & value \\
        \hline
        hidden size & 1024 \\
        batch size & 8 \\
        gradient accumulation & 16 \\
        epoch & 10 \\
        learning rate & 1e-5 \\
        weight decay rate & 0.1 \\
        warming up ratio & 0.06 \\
        optimizer & adam \\
        doc stride & 150 \\
        maximum pieces & 384 \\
        \specialrule{.1em}{.05em}{.05em}
    \end{tabular}
    \caption{The hyperparameters}
    \label{hyper}
\end{table}

\vspace{-0.2cm}
\paragraph{Metrics}
Though ROPES was released using both exact match (EM) and F1 as metrics, we only report EM here, as F1 has been shown to correlate poorly with human judgments on ROPES~\cite{chen-etal-2019-evaluating}.  F1 assumes that answers that share many overlapping words are likely similar; while this is largely true on SQuAD~\cite{rajpurkar2016squad}, where this particular F1 score was introduced, it is not true on ROPES, where things like \emph{Village A} and \emph{Village B} are both plausible answers to a question.  All the systems are trained in three runs with different random seeds, and we post the average performance over the three runs.

\subsection{Results}
Table~\ref{results} shows the performance of the three systems. The multi-step system and multi-step reranker outperform the baseline system with 8.1\% and 11.7\% absolute EM accuracy on dev set, respectively, and with 2.4\% and 2.0\% EM accuracy on test set, respectively, showing that with multi-step inference, the system can achieve improvements. With the ensemble, the multi-step reranker performs best on dev and test sets.

As can be seen, the improvement of our model on the dev set is quite large.  While performance is also better on the official test set, the gap is not nearly so large.  To understand whether this was due to overfitting to the dev set or to the distributional shift mentioned in Section~\ref{bias}, Table~\ref{results} also shows the results on dev-test, our split that treats the official dev set as a held-out test set.  Here, we still see large gains of 7.2\% EM from our model, suggesting that it is indeed a distributional shift and not overfitting that is the cause of the difference in performance between the original dev and test sets.  Properly handling the distributional shift in the ROPES test set is an interesting challenge for future work.

\begin{table}[!t]
    \centering
    \begin{tabular}{lcc|c}
    \toprule 
          Model     & {Dev} & {Test} & Dev-test \\
      \midrule
      Baseline & 59.7 & 55.4 & 56.2\\
      Multi-step & 67.8 & 57.8 & 61.6\\
      Multi-step reranker & 71.4 & 57.4 & 63.4\\
      ~~~~+ensemble & 73.3 & 58.8 & 65.2\\
      \bottomrule 
    \end{tabular}
    \caption{The exact match scores by three systems. For the first two columns, we performed model selection on dev; for the third column, we performed model selection on a separate train-dev set.}
    \label{results}
\end{table}

\subsection{Analysis and Discussion}
\label{analysis}
We conduct detailed analysis in this section, studying (1) the impact of various components of our model, (2) the gap between results on development and test set, (3) the strategy for sampling candidates for the reranker, and (4) the errors that the models cannot cover. 

\paragraph{Ablation Study}
We perform an ablation study on the multi-step system and the multi-step reranker. Table~\ref{ablation} shows the results on dev set by various ablated systems. The performances of two systems drop down without any one module due to the property of the chained reasoning. The performance of the multi-step system without Q \textsc{Select}  or B \textsc{Chain} drops (around) more than that of the multi-step system without B \textsc{Select} or S \textsc{Chain} (around -2.1\% EM ). So Q \textsc{Select} module and B \textsc{Chain} play relatively more important roles. The performance of the multi-step reranker without Q \textsc{Select}, B \textsc{Select} or S \textsc{Chain} drops (around -5.9\% EM) more than that of the multi-step reranker without B \textsc{Chain} (-3.7\% EM). 


\begin{table}[!t]
    \centering
    \begin{tabular}{l|l}
    \toprule 
      Model   &  \multicolumn{1}{c}{EM} \\
    \midrule
    Multi-step  & 67.8 \\
    ~~~~w/o Q \textsc{Select} & 62.8 (-5.0)  \\
    ~~~~w/o B \textsc{Chain} & 62.3 (-5.5) \\
    ~~~~w/o B \textsc{Select} & 65.9 (-1.9) \\
    ~~~~w/o S \textsc{Chain} & 65.5 (-2.3) \\
    Multi-step reranker  & 71.4 \\
    ~~~~w/o Q \textsc{Select} & 65.8 (-5.6) \\
    ~~~~w/o B \textsc{Chain} & 67.7 (-3.7) \\
    ~~~~w/o B \textsc{Select} & 64.9 (-6.5) \\
    ~~~~w/o S \textsc{Chain} & 65.7 (-5.7) \\
    \bottomrule
    \end{tabular}
    \caption{The ablation results on development. Q \textsc{Select} denotes the question \textsc{Select} module; B \textsc{Chain} denotes the \textsc{Chain} module applied on the background and the question; B \textsc{Select} denotes the background \textsc{select} module; S \textsc{Chain} denotes the \textsc{Chain} module applied on the situation and the previous chained reasoning.
    }
    \label{ablation}
\end{table}
\paragraph{Answer Types}
We break down the overall accuracy by answer type, which is shown in Table~\ref{types-score}. All three systems perform substantially better on NP, ADJP, and ADVP questions than on VP questions. The main reason is that the VP questions are associated with complex and long answers, e.g., \textit{acquire it from other organisms} or \textit{make their own glucose}.  The major improvements happen on answering NP and ADVP questions, which explains the gap between the scores on the development set, with a large amount of NP questions, and the test set, with relatively more VP questions. The analysis can inspire the future work of investigating the specific inference programs for specific-type questions.

\begin{table}[!t]
    \centering
    \begin{tabular}{@{~}l@{~}@{~}c@{~}c@{~}c@{~}c@{~}@{~}c@{~}}
    \toprule 
       Model  &  NP & VP & ADJP & ADVP & avg \\
    \midrule
    Baseline & 60.0 & 38.1 & 60.4 & 62.7 & 53.03 \\
    Multi-step & 68.8 & 39.7 & 61.3 & 72.6 & 58.65 \\
    Multi-step reranker & 71.8 & 38.1 & 63.8 & 75.0 & 60.52\\
    ~~~~+ensemble & 75.0 & 42.9 & 61.3 & 78.6 & 62.75\\
    \bottomrule 
    \end{tabular}
    \caption{The exact match accuracy of most four frequent question types in test dataset. avg is the weighted accuracy in terms of frequency of the four kinds of questions.}
    \label{types-score}
\end{table}

\paragraph{Candidate Sampling}
In order to train the reranker, we need training data with high-diversity candidates. However, a well-trained model does not generate similar candidates for the training set to what it generates for the dev and test sets, due to overfitting to the training set.  In order to get useful candidates for the training set, we need a model that was not trained on the data that it generates candidates for. We investigate four strategies based on cross-validation to generate training data candidates: 10-fold, 5-fold, 2-fold and 3-turn. With the $k$-fold method, the training data is partitioned into $k$ parts, and $(k-1)$ parts are used to train a model that generates candidates answers for the remaining part. With the $k$-turn method, the training data is partitioned into $k$ parts, and the $i$th part is used to train a model that generates candidate answers for $(i+1)$th part. 

Table~\ref{strategy} shows the average accuracy on training data. The accuracy on training data generated by $k$-fold self-sampling method is very high, and they are not consistent with the dev and test set. The accuracy on training data generated by the 3-turn self-sampling method is most similar to the accuracy on dev set (59.7\% EM) and test set (55.4\% EM) by the baseline system. We adopt the 3-turn self-sampling method for our experiments.

\begin{table}[!t]
    \centering
    \begin{tabular}{c|c}
    \specialrule{.1em}{.05em}{.05em} 
         &  EM \\
    \hline
     10-fold & 84.1 \\
     5-fold & 82.4 \\
     2-fold & 75.9 \\
     3-turn & 59.9 \\
     \specialrule{.1em}{.05em}{.05em} 
    \end{tabular}
    \caption{The average accuracy on training data for the multi-step reranker.}
    \label{strategy}
\end{table}

Table~\ref{topk} shows the oracle of top $k$ candidates on train, development and test set. Because oracle scores are the upper bound of the reranker, there is a trade-off that the upper bound is lower as fewer candidates are sampled, while the noise increases as more incorrect candidates are sampled. We found that top 3 provides a good trade-off for the reranker on the development set, giving a large jump over just two candidates, and this is what we used during our main experiments.
\begin{table}[!t]
    \centering
    \begin{tabular}{c|ccc}
    \specialrule{.1em}{.05em}{.05em}
        $k$ & {train} & {dev} &{test} \\
        \hline
        1 & 59.9 & 59.7 & 55.4  \\
        2 & 81.4 & 64.8 & 61.9  \\
        3 & 92.0 & 97.4 & 80.2 \\
        4 & 93.8 & 98.3 & 83.6 \\
        5 & 94.9 & 98.7 & 85.9 \\
        10 & 96.1 & 99.4 & 88.5 \\ 
        \specialrule{.1em}{.05em}{.05em}
    \end{tabular}
    \caption{The oracle scores for top $k$ candidates.}
    \label{topk}
\end{table}

\paragraph{Error Analysis and Future Work}
We analyze some errors that our proposed model made, aiming to discover the questions that our model could not cover.
Table~\ref{case} shows some questions that our proposed model gives incorrect answers. The questions require model to get the numeric information from the passage, and then compare the numeric relation (e.g. larger, smaller and equal) and target the effect of the relation in the background passage, where positive correlation between the prices and the sold number in example 1, positive correlation between the tolerance degree and usage times in example 2 and negative correlation between the crash rate the the number of cyclists in example 3. It seems that the model is not sensitive to the numeric information and their reasonings.

Also, the situations give more than two entities with their related information, and although the questions narrow down the multiple choices to two choices, the systems are still distracted by these question-irrelevant entities. The distraction comes from the difficulty of associating the relevant information with the correct entities. Future work can be motivated by the discovery to design more modules to deal with this phenomenon.


\begin{table}[!t]
    \centering
    \begin{tabular}{p{7.2cm}}
    \specialrule{.1em}{.05em}{.05em}
    Example 1 \\
    \specialrule{.1em}{.05em}{.05em}
    \textbf{Background:} {\small ... For many of the works, the price goes up as the edition sells out...}\\
    \hline
  \textbf{Situation:} {\small ...By the end of the week, they started to sell out. There were only 2 of the Mona Lisa,...,120 of The Kiss, 150 of The Arnolfini Portrait...} \\
    \hline
    \textbf{Question:} {\small Which limited edition most likely had it's price increased: The Kiss or Mona Lisa ?} \\
    \textbf{Answer:} {\small The Kiss} \\
    \hline
    Ours:{\small Mona Lisa} \\
    \specialrule{.1em}{.05em}{.05em}
    \specialrule{.1em}{.05em}{.05em}
    Example 2 \\
  \specialrule{.1em}{.05em}{.05em}
  \textbf{Background:} {\small ...The tolerance for a drug goes up as one continues to use it after having a positive experience with a certain amount the first time...}\\
    \hline
  \textbf{Situation:} {\small ... Chris used it 12 times,...,Jimmy used it 42 times, Antonio used it 52 times, Danny used it 62 times, ...} \\
    \hline
    \textbf{Question:} {\small Who has a higher tolerance for roach: Jimmy or Antonio ?} \\
    \textbf{Answer:} {\small Antonio} \\
    \hline
    Ours: {\small Jimmy}\\
    \specialrule{.1em}{.05em}{.05em}
    \specialrule{.1em}{.05em}{.05em}
    Example 3 \\
  \specialrule{.1em}{.05em}{.05em}
  \textbf{Background:} {\small ... That is to say, the crash rate per cyclist goes down as the cycle volume increases... }\\
    \hline
  \textbf{Situation:} {\small ...Day 1 had 500 cyclists left.  Day 2 had 400 cyclists left. Day 3 had 300 cyclists left. Day 4 had 200 cyclists left....} \\
    \hline
    \textbf{Question:} {\small What day had a lower crash rate per cyclist: Day 1 or Day 2 ?} \\
    \textbf{Answer:} {\small Day 1} \\
    \hline
    Ours: {\small Day 2} \\
    \specialrule{.1em}{.05em}{.05em}
    \end{tabular}
    \caption{The examples of the answers to the questions by the multi-step reranker.}
    \label{case}
\end{table}

\section{Related Work}
\paragraph{Neural Module Networks} were originally proposed for visual question answering tasks~\cite{nmn1,andreas-etal-2016-learning}, and recently have been used on several reading comprehension tasks \cite{jiang-etal-2019-explore,jiang-bansal-2019-self,Gupta2020Neural}, where they specialize the module functions such as \textsc{Find} and \textsc{Compare} to retrieve the relevant entities with or without supervised signals for HotpotQA~\cite{yang2018hotpotqa} or DROP~\cite{dua2019drop}.  As ROPES is quite different from these datasets, the modules that we choose to use are also different, focusing on chained inference.

\paragraph{Multi-Hop Reasoning} There are several datasets constructed for multi-hop reasoning e.g. \textsc{HotpotQA} \cite{yang2018hotpotqa,jiang-etal-2019-explore,jiang-bansal-2019-self,min-etal-2019-multi,feldman2019multi}, \textsc{QAngaroo} \cite{welbl2018constructing,chen2019multi,zhuang-wang-2019-token,tu-etal-2019-multi} and \textsc{WikiHop} \cite{welbl2018constructing,song2018exploring,das2019multi,asai2019learning} which aims to get the answer across the documents. The term ``multi-hop" reasoning on these datasets is similar to relative information retrieval, where one entity is bridged to another entity with one hop. Differently, the multi-step reasoning on ROPES aims to do reasoning over the effects of a passage (background and situation passage) and then give the answer to the question in the specific situation, without retrieval on the background passage.

\paragraph{Models beyond Pre-trained Transformer} As the emergence of fully pre-trained transformer \cite{peters2018deep,devlin2019bert,liu2019roberta,radford2019language,dai-etal-2019-transformer,yang2019xlnet}, most of NLP benchmarks got new state-of-the-art results by the models built beyond the pre-trained transformer on specific tasks (e.g. syntactic parsing, semantic parsing and GLUE) \cite{wang2018glue,Kitaev-2018-SelfAttentive,zhang-etal-2019-amr,tsai2019small}. Our work is in the same line to adopt the advantages of pre-trained transformer, which has already collected contextualized word representation from a large amount of data.

\section{Conclusion}
We propose a multi-step reading comprehension model that performs chained inference over natural language text. We have demonstrated that our model substantially outperforms prior work on ROPES, a challenging new reading comprehension dataset. We have additionally presented some analysis of ROPES that should inform future work on this dataset. While our model is not a neural module network, as our model uses a single fixed layout instead of different layouts per question, we believe there are enough similarities that future work could explore combining our modules with those used in other neural module networks over text, leading to a single model that could perform the necessary reasoning for multiple different datasets.

\section*{Acknowledgments}
We thank the anonymous reviewers for their feedback. We gratefully acknowledge the support of the European Research Council (Lapata, Liu; award number 681760), the EU H2020 project SUMMA (Cohen, Liu; grant agreement 688139) and Bloomberg (Cohen, Liu).

\bibliography{acl2020}
\bibliographystyle{acl_natbib}

\end{document}